# Eliminating Search Intent Bias in Learning to Rank

Yingcheng Sun
Case Western Reserve University
Cleveland, OH, USA
yxs489@case.edu

Richard Kolacinski
Case Western Reserve University
Cleveland, OH, USA
rmk4@case.edu

Kenneth A. Loparo
Case Western Reserve University
Cleveland, OH, USA
kal4@case.edu

*Abstract*—Click-through data has proven to be a valuable resource for improving search-ranking quality. Search engines can easily collect click data, but biases introduced in the data can make it difficult to use the data effectively. In order to measure the effects of biases, many click models have been proposed in the literature. However, none of the models can explain the observation that users with different search intent (e.g., informational, navigational, etc.) have different click behaviors. In this paper, we study how differences in user search intent can influence click activities and determined that there exists a bias between user search intent and the relevance of the document relevance. Based on this observation, we propose a search intent bias hypothesis that can be applied to most existing click models to improve their ability to learn unbiased relevance. Experimental results demonstrate that after adopting the search intent hypothesis, click models can better interpret user clicks and substantially improve retrieval performance.

*Keywords—Search Intent; Click Bias; Learning to Rank*

## I. Introduction

Learning to rank (LTR) is an important approach to creating ranking models in information retrieval by using training data to train ranking functions and testing data to evaluate their performance. Search engine click through logs are widely used to construct the training and testing datasets for LTR for several reasons; the data is inexpensive to collect, easy to maintain and can capture the temporal patterns actual user preferences. Despite these clear benefits, LTR from click through data is challenging due to the existence of noises and behavior biases in the click data.

Previous studies have shown that clicking behaviors are biased by aspects such as "position" [1] (a document appearing in a higher position is more likely to attract user clicks even though it is not as relevant as other documents in lower positions), "presentation" [2] (different search results display styles that will influence users' attention allocation strategies), "domain" [3] (a user's propensity to believe that a page is more relevant just because it comes from a particular domain.), and so on. To address these problems, researchers have proposed a number of click models to describe a user's practical browsing behavior and to obtain an unbiased estimate of result relevance [4, 5, 47].

To the best of our knowledge, user search intent and its influence on click behaviors has not been well studied in the existing literature. The search intent of users can be classified into three categories according to user goals: informational, navigational and transactional [6]. In navigational queries, the user is interested in reaching a specific web site like "GitHub", "Harvard Business School", etc. In informational queries, the user does not have a particular page in mind and intends to learn more about a specific topic, such as ''amazon forest'', "American civil war", etc. In transactional queries, the user's objective is to obtain a specific resource, and examples are the "software downloads", "purchasing airline tickets", etc. A phenomenon uncovered by Guan and Cutrell [7] is that click behavior of a user can vary depending on their search intent, and observed that people clicked much less frequently on lower-ranked results for informational tasks than navigational tasks. Because their comparisons only included click behaviors between two types of search intents, which, we performed experiments and studied click activities among all the three types of search intents. Experimental results show that users with navigational or transactional search intent are more likely to click the target results than those with informational search intent, especially when the targets are ranked in a lower position. For example, given the transactional search query "Introduction to Algorithms pdf", users are patient enough to find the target link in the result list even it is in a lower position. When the search query is informational like "sorting algorithms for big data", however, users have a tendency to click the top results whether or not the specific target is relevant.

We thus propose search intent bias to explain how user clicks are affected by search intent in addition to relevance and positon bias. We argue that users with different search intents may have different click behaviors, and this bias should be modeled by characterizing search intent diversity. After eliminating search intent bias in learning to rank, the analytical power of the examination hypothesis will be enhanced and the user clicks in search engine logs will be interpreted better. Also, our proposed search intent bias is straightforward and can be easily applied to most of existing click models to improve their capacities in learning unbiased relevance.

In this paper, we first introduce the related work and background for our model, and then we present effective features for recognizing the user intent behind search queries, and use these information to automatically classify search intents. Next, we show the difference of users' clicking behaviors with different search intents, and propose our search intent bias model and its inference method. We estimate the click model using search intent bias data and the Expectation-Maximization (EM) algorithm. In the experimental section, we introduce the datasets, comparison methods and evaluation metrics, as well as the experimental results. The results demonstrate that click models extended by search intent bias can better interpret user clicks. We then conclude our work and explore possible future research in the last section.

## II. RELATED RESEARCH AND BACKGROUND

As discussed in the first section, click through data cannot be directly used for learning to rank due to the biases it brings. To leverage the full power of click through data, researchers have attempted to remove the effect of user bias in the training of ranking models. One such effort is the development of click models. Click models make hypotheses about user browsing behaviors and estimate the true (unbiased) relevance feedback by optimizing the likelihood of the observed user clicks. Ranking models are then trained with the estimated relevance signals so that the overall system is unbiased [8].

Most click models follow the examination hypothesis [9]: a document being clicked should satisfy two conditions: it is examined and relevant. Assume the *i-th* document is denoted by $d_{\pi(i)}$ and whether it is clicked is denoted by $C_i$. $C_i$ is a binary variable. $C_i = 1$ represents that the document is clicked and $C_i = 0$ represents that it is not clicked. Similarly, whether the document $d_{\pi(i)}$ is examined and whether it is relevant are respectively represented by the binary variables $E_i$ and $R_i$. The examination hypothesis can be formulated as

$$E_i = 1, R_i = 1 \Leftrightarrow C_i = 1 \quad (1)$$

where $R_i$ and $C_i$ are independent of each other. The value of $C_i$ is observable from search sessions and the value of $E_i$ and $R_i$ are hidden. We then use $\Pr(C_i = 1)$ as the click through rate of the *i*-th document, $\Pr(E_i = 1)$ is the probability of examining the *i*-th document, and $\Pr(R_i = 1)$ is the relevance of the *i*-th document. We use the parameter $r_{\pi(i)}$ to represent the document relevance as

$$P(R_i = 1) = r_{\pi(i)} \quad (2)$$

Formula (1) can then be reformulated in a probabilistic way:

$$P(C_i = 1 | E_i = 1, R_i = 1) = r_{\pi(i)} \quad (3)$$
$$P(C_i = 1 | E_i = 0) = 0 \quad (4)$$
$$P(C_i = 1 | R_i = 0) = 0 \quad (5)$$

After summation over $R_i$, this hypothesis can be simplified as

$$P(C_i = 1 | E_i = 1) = r_{\pi(i)} \quad (6)$$
$$P(C_i = 1 | E_i = 0) = 0 \quad (7)$$

Thus, the probability of a document being clicked is determined as follows:

$$P(C_i = 1) = \sum_{e \in (0,1)} P(E_i = e) P(C_i = 1 | E_i = e)$$
$$= P(E_i = 1) P(C_i = 1 | E_i = 1)$$

A series of click models have different implementations of examination hypothesis. For example, the cascade model [1] assumes a user will examine the search results sequentially without skipping from top to bottom until she clicks a result. Therefore, a document is examined only if all previous documents are examined. Dependency Click Model (DCM) [10] extends the cascade model in order to model user interactions within multi-click sessions. DCM assumes that a user may have a certain probability of examining the next document after clicking the current document, and this probability is influenced by the ranking position of the result. Dynamic Bayesian Network (DBN) [4] extends the cascade model by assuming that the examination probability also depends on the clicks and the relevance of previous documents. DBN uses a satisfaction parameter $S_i$, which states if the user is satisfied with the clicked document, she will not examine the next document. Otherwise, there is a probability $\gamma$ that the user will continue her search.

$$P(S_i = 1 | C_i = 0) = 0 \quad (8)$$
$$P(S_i = 1 | C_i = 1) = s_{\pi(i)} \quad (9)$$
$$P(E_{i+1} = 1 | S_i = 1) = 0 \quad (10)$$
$$P(E_{i+1} = 1 | E_i = 1, S_i = 1) = \gamma \quad (11)$$

Subsequently, the User Browsing Model (UBM) [5] further refines the examination hypothesis by assuming that the event of a document being examined depends on both the preceding click position and the distance between the preceding click position and the current one. It assumes that the examination event $E_i$ depends not only on the position $i$ but also on the previous clicked position $l_i$ in the same query session, where $l_i = max\{j \in \{1, \cdots, i-1\} | C_j = 1\}$, and $l_i = 0$ means no preceding clicks. Global parameters $\beta_{l(i),i}$ measure the transition probability from position $l_i$ to position $i$, and $C_{i:j} = 0$ is an abbreviation for $C_i = C_{i+1} = \cdots = C_j = 0$:

$$P(E_i = 1 | C_{1:i-1} = 0) = \beta_{0, i} \quad (12)$$
$$P(E_i = 1 | C_{l(i)} = 1, C_{l(i)+1 : i-1} = 0) = \beta_{l(i), i} \quad (13)$$
$$P(C_i = 1 | E_i = 0) = 0 \quad (14)$$
$$P(C_i = 1 | E_i = 1) = r_{\pi(i)} \quad (15)$$

Other models like Mobile Click Model (MCM) [11], Personalized Click Model (PCM) [12], and etc. try to capture other type of bias to extract more accurate signals from click data.

Recently, with the development of adversarial training [44, 45] and context based machine learning technologies [30, 31, 32, 33, 34, 48, 49], unbiased LTR has been actively studied to learn a ranking function directly from biased click data based on the counterfactual inference framework [13]. This unbiased learning-to-rank framework treats click bias as a counterfactual effect and debiases user feedback by weighting each click with their Inverse Propensity Scoring (IPS). Under this framework, Wang et al. [14] proposed a result randomization method that randomly shuffles the top n results and uses the average CTR at each position as the propensity. Agarwal et al. [15] propose a noise aware Position-Based Model (TrustPBM) to address click noise for unbiased LTR. A modification to the existing IPS estimator is proposed to take into account the bias due to the context in which a document was presented to the user [16]. Wang et al. [12] propose a regression based EM algorithm that is based on a position bias click model to handle highly sparse clicks in personal search without relying on randomization. Related research on this topic is also discussed in [35-43].

All these studies try to interpret the collected click through data to minimize various bias effects or to learn from biased click logs directly, but none of them addressed the problem that different types of search intents will lead to different click behaviors. Hu et al. [17] discuss the hypothesis that users with different search intents may submit the same query but expect different search results. Thought with similar topics (intent bias), we focus on multiple types of search queries instead of multi-meanings of a query thus with different research perspectives.

III. MODELING SEARCH INTENT BIAS

In this section, we first list the features used for search intent classification, and then show the difference of users' clicking behaviors with different search intents, and propose our search intent bias and its inference method in the end.

*A. Search Intent Classification*

Before studying the influence of different types of user search intents lead to user click behaviors, we first need to detect the users' goals behind their queries and classify their search intents into different classes. Various intent taxonomies have been explored in the past years [18, 19]. In this paper, we adopt the most well-known taxonomy proposed by Broder [6] classifies search queries into informational, navigational and transactional.

Similar to most natural language processing tasks, there are two main approaches to identifying query intent: rule-based and statistical methods [28]. The rule-based systems use predefined rules to match new queries to their intents. While these systems are usually precise, their coverage is low. In addition, the rules need to be carefully engineered by human experts; designing a new rule for the system might take a few hours. Scaling up these systems to a large number of intents is difficult and needs a huge amount of human effort, so we use statistical methods. Different from documents, queries are usually shorter and lack sufficient context to be classified [27]. We thus need enough features to represent a query. Table 1 lists three types of features we adopted in this study which are grouped according to how they are extracted.

TABLE I. FEATURES USED FOR DETECTING SEARCH INTENTS.

| | Search Query Features | |
|---|---|---|
| Linguistic | Bag-of-Words (BoW) | a web query as a term frequency vector |
| | Query length | the number of tokens within the query |
| | POS tagging categories | computed by the caseless models |
| | Named entity classes | computed by the caseless models |
| | Dependency trees typed relationships | given by the caseless models |
| Context | Anchor Text Frequency Distribution | the distribution of occurrences of the query in the anchor texts of the documents in the query answer set |
| | URL Feature | The match between query and domain |
| | WordNet | Semantic relations between words. |
| | Explicit semantic analysis (ESA) | the meaning of any text as a weighted vector composed of the top-k related Wikipedia-based concepts |
| Click-through | Click Ratio | The ratio of number of clicks on a particular URL to the total number of clicks |
| | N Clicks Satisfied (nCS) | the number of sessions of a query that register less than n selections |
| | n Results Satisfied (nRS) | the number of sessions of a query q that register selections only in the top-n results of the answer list of q |

We choose informative, independent and simple features for query classification. Linguistic features [20] express lexical attributes of queries, distilled from the output computed by the Stanford NLP tools. Web queries are frequently lack of context, therefore we use context features to expand queries with semantic information. Extracted from click through data, click-through features model user click preference [27]. We pick up few of them to explain in details.

POS tagging categories were computed by the caseless models implemented by Stanford Tagger[1]. Named entity classes were supplied by the caseless models implemented by Stanford Named Entity Recognizer [2]. Dependency trees typed relationships were given by the caseless models implemented by Stanford lexicalized dependency Parser[3]. URL features measures the match between query and URL address, named *urlmr*, which is defined as:

$$urlmr = l(p)/l(u)$$

Where *l(p)* is the length of the longest substring *p* of the query that presents in the *URL* and *l(u)* is the length of the URL *u* [27]. Explicit semantic analysis (ESA) represents the meaning of any text as a weighted vector composed of the top-k related Wikipedia-based concepts [25]. This semantic space is particularly useful for short documents [26]. We profited from the implementation provided by ESAlib[4]. For the click ratio, let $n_k^i$ denote the number of clicks on URL k for query *i* and total number of clicks:

$$n^i = \sum_k n_k^i$$

The click ratio is the number of clicks on a particular URL *K* for query *i* to the total number of clicks for this query. Which has the form:

$$CR(i, K) = \frac{n_K^i}{n^i}$$

N Clicks Satisfied (nCS) evidence [21] is extracted from the number of user clicks for a particular query. It is based on the following assumption: while performing a navigational type search request, user tend to click a small number of URLs in the result list. Supposing one web search user has a navigational goal, he has a fixed search target in mind and would like to find just that target URL and corresponding snippet in the result list. So it is impossible for him to click a number of URLs which are not the target page unless there exists cheating pages. According to this assumption, a query type can be judged by the number of URLs which the user clicks:

$$nCS\ (Query\ q)\ =\ \frac{\#(Session\ of\ q\ that\ involves\ less\ than\ n\ clicks)}{\#(Session\ of\ q)}$$

Top n Results Satisfied (nRS) evidence [21] is extracted from the clicked URL's rank information. It is based on the following assumption: while performing a navigational type search request, user tend to click only the first few URLs in the result list. This assumption is related to the fact that navigational type queries have a much higher retrieval performance than informational/transactional ones, so we can judge a query type by whether the user clicks other URLs besides the first *n* ones. Top *n* Results Satisfied (nRS) feature developed from this idea is defined as:

$$nRS\ (Query\ q)\ =\ \frac{\#(Session\ of\ q\ that\ involves\ clicks\ only\ on\ top\ n\ results)\ )}{\#(Session\ of\ q)}$$

[4] http://lukas.zilka.me/esalib/

## B. Click Behavior Study

After getting the category labels of search queries, we are interested in how users' search behaviors vary with different types of search tasks. In the experiment of Guan et al. [7], they examined users' click behaviors given two types of search tasks: navigational and informational, and found that when the targets locate in relatively low positions in the search results, people click more frequently on the top results and hit less successfully on the target for informational tasks than navigational tasks. In order to formalize and model this potential search intent bias, we study all types of search intent in this paper and make sure each query can be "covered" by any one of intent category.

We randomly extract a group of queries for each of the three search intents (informational, navigational and transactional) from click logs collected from two commercial search engine: AOL[5] and Sogou[6]. The search logs in both datasets recorded the positon/ rank on which the user clicked in the search result list given a query. However, they are not labeled by the search intent and the relevance (whether the clicked result is the target or not), so we need to annotate them manually. First, URLs that are broken or outdated are filtered out, and the queries by users who are looking for pornographic material are removed, too. Next, we annotate the relevance of each clicked URL to query, and the search intent of each query. Figueroa et al. [20] provide the annotation of search intents for AOL dataset, so we only need to work on the Sogou dataset.

The canonical definition [6, 19] is not appropriate for annotation as guideline. For example, if a user issues query "amazon", he/she mainly wants to visit "amazon.com", so this query is navigational because the user's goal is to reach a particular website. This definition, however, is rather subjective and not easy to formalize. In this paper, we extend the definition of navigational query to a more general case: a query is navigational if it has one and only one perfect site in the result set corresponding to this query [27]. A site is considered as perfect if it contains complete information about the query and lacks nothing essential. In this definition, navigational query must have a corresponding result page that conveys perfectness, uniqueness, and authority. Unlike Broder's definition, our definition does not require the user to have a site in mind. This makes data labeling more objective and practical. For example, when a user issues a query "Fulton, NY", it is not clear if the user knows the Web-site "www.fultoncountyny.org". However, this Web-site has an unique authority and perfect content for this query and therefore the query "Fulton, NY" is labeled as a navigational query. For transactional queries, we use the definition introduced in [29]: a query is transactional when it has cue words related to "file", "video", "music", "picture" and "travel". All the other queries are considered informational. For an informational query, typically there exist multiple excellent websites corresponding to the query that users are willing to explore.

After the annotation, we calculate the click distribution for each type of search intent. Fig. 1 shows the visualized statistical result. In Fig.1, the number inside the bubble refers the click rate in percentage. For some queries, the sum of click rates on results is less than 1. It is because few users issued a new query without clicking any of the results. For example, the sum of click rates with the target in position 4 in navigational search group is 97%, meaning that 3% submissions on this query had no click records.

We can see that for navigational tasks, people had the highest click accuracy rate when the target was in the first 2 positions (96%, 92%). With the target at position 4, 5, 7, and 8, click accuracy dropped to 67% or less, and the click through rate on positon 1 and 2 are 25.5% and 14.5% on average, that are very close to transactional tasks (28.8%, 14.5%), but less than informational tasks (30.5%, 18.8%). For informational tasks, the effect of target position on click accuracy was also much more dramatic, and participants correctly selected the target less than 20% of the time when the target was below position 2.

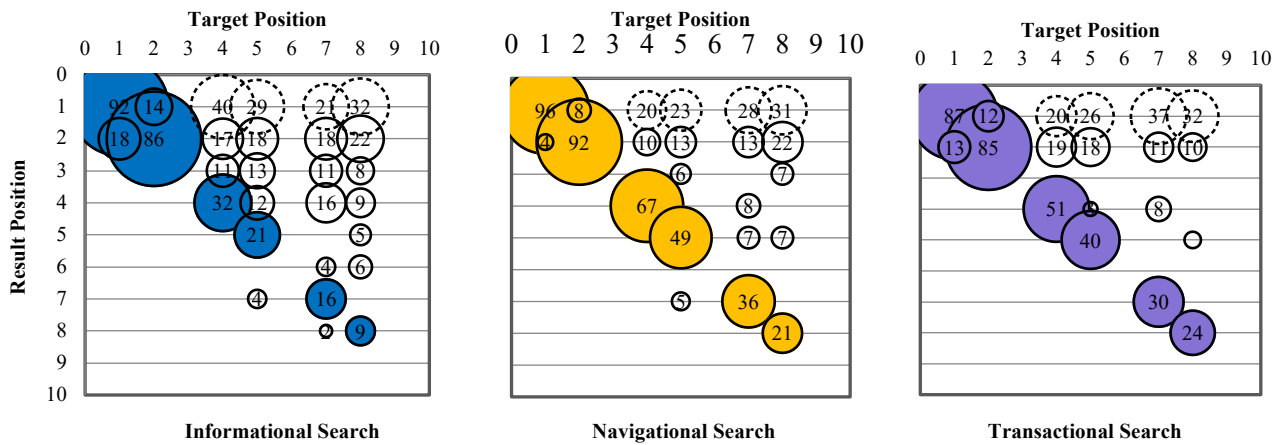

Figure. 1. Chance of clicking on search results break down with target position. The numbers inside the bubble indicate the chance (in %) that the result was clicked (e.g. when the target position is 1 for informational search, 92% of users clicked on result 1, which is the target result.) The shadowed bubbles indicate the target results. The bubble with a dashed border indicates the first result. Bigger bubble indicates a larger probability of clicking at the result at its particular position, which is also shown with a number inside the bubble.

---
[5] https://archive.org/details/AOL_search_data_leak_2006
[6] https://www.sogou.com/labs/resource/q.php

This suggests that users trust the search engine more for informational search or invest less scrutiny in judging the results with higher rankings. Eventually they are more likely to choose the top few results to try them out than navigational or transactional search intents. Such results motivate us to propose a new click model to cope with the corresponding click bias caused by user search intent.

*C. Model Intent Bias*

We formally introduce the search intent bias and as well as how we incorporate them into click models in this subsection, and show the graphical representation of the click model extended with search intent bias. We first describe the search intent bias as follows.

We follow the examination hypothesis that most click models follows: *A document is clicked if and only if it is both examined and relevant*, which can be formulated as

$$E_i = 1, R_i = 1 \Leftrightarrow C_i = 1 \quad (16)$$

Under this assumption, the search intent bias can be formulated as:

$$P(E_i = 1, S_t) = \gamma_{\pi(i),t} \quad (17)$$
$$P(C_i = 1 | E_i = 0, S_t) = 0 \quad (18)$$
$$P(C_i = 1 | E_i = 1, S_t) = r_{\pi(i),t} \quad (19)$$

$P(C_i = 1) = P(E_i = 1, S_t) \cdot P(C_i = 1 | E_i = 1, S_t) = \gamma_{\pi(i),t} \cdot r_{\pi(i),t}$ (20) where $\gamma_{\pi(i),t}$ is the examination parameter representing the position bias for document $d_{\pi(i)}$ and $r_{\pi(i),t}$ is the relevance of the document $d_{\pi(i)}$ in the returning results with query intent $t$. The above model is an extension to classical position-based model (PBM) by adding search intent bias factor $S_t$. Users show different click behaviors under different search intents, so $S_t$ acts as an annotation variable to label the position bias and document relevance with various search intents. Thus the search intent bias will be measured by the position bias variable $\gamma_{\pi(i),t}$ and document relevance variable $r_{\pi(i),t}$. We present it as a graphical model in Fig. 2. We assume that the search intent is already known after query classification, so $S_t$ is an observed variable in this model.

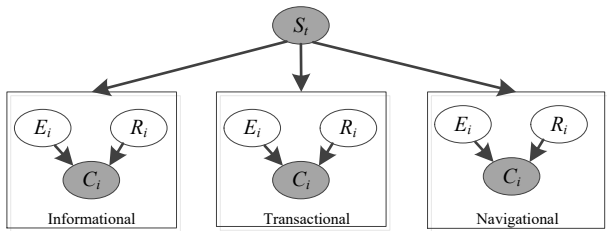

Figure. 2. Graphical representation of the extended position-based model. Solid node represents observed variable and white node represents hidden variable.

Three sets of variables in our model need to be inferred, but each set only contains two unobserved variables which is exactly the same as PBM, so Expectation-Maximization algorithm can be used to perform the inference process.

For all previous click models based on the examination hypothesis, the switch from the examination hypothesis to the search intent - aware model is quite simple. Actually, we only need to label the original position bias variable and document relevance variable with search intent type without changing any other specifications. When an unbiased model is constructed, we estimate the value of $\gamma_{\pi(i),t}$ for each session. After all of the $\gamma_{\pi(i),t}$ are known, then other parameters of the click model (such as relevance) can be learned. However, since the estimation of $\gamma_{\pi(i),t}$ relies on learning the results of other parameters, the entire inference process has deadlocks. To avoid this problem, we adopt an iterative inference. Every iteration consists of two phases. In Phase A, we learn the click model parameter $\Theta$ based on the estimated values of $\gamma_{\pi(i),t}$ of the last iteration. In Phase B, we estimate the value of $\gamma_{\pi(i),t}$ for each session based on the parameters $\Theta$ learned in Phase A. Here, the likelihood function that we want to maximize is the conditional probability that the actual click events of this session occur under the specification of the click model, with $\gamma_{\pi(i),t}$ being treated as the condition. Phase A and Phase B should be executed alternatively and iteratively until all parameters converge.

## IV. EXPERIMENTS

In this section, we evaluate our proposed hypothesis of search intent bias, and demonstrate the effectiveness after extending traditional click models to intent-aware models. We conduct a series of experiments on large scale search logs to answer the following research questions:

- RQ1: Does intent aware models have better click prediction ability than the baseline models?

- RQ2: Can intent aware models provide better relevance estimations of search results than the baseline models?

*A. Dataset*

The search engine logs used to train and evaluate click models are collected from a commercial search engine AOL in the U.S. market during March and May in 2006, and a popular Chinese search engine Sogou in June in 2008. AOL data set consists of about 21 million of search queries prompted by approximately 650,000 users, and Sogou data set contains about 86 million search queries. Each instance has a user id, timestamp, search query string, the rank and the URL of the clicked results. Each line of the search log includes five items:

– AnonID - an anonymous user ID number.

– Query - the query issued by the user, case shifted with most punctuation removed.

– QueryTime - the time at which the query was submitted for search.

– ItemRank - if the user clicked on a search result, the rank of the item on which they clicked is listed.

– ClickURL - if the user clicked on a search result, the domain portion of the URL in the clicked result is listed.

Each line in the data represents one of two types of events: a query that was NOT followed by the user clicking on a result item or a click through on an item in the result list returned from a query.

Figueroa [20] annotated 60,000 queries of AOL data set with search intents [1], that are used in our experiment. We also annoated 10,000 queries from Sogou dataset both automatically by discriminative keywords [20] and manually by three annotators. The lowest disagreement rate was of 2%, while the highest 15%. The average disagreement rate was 8% (Std. dev. ±4.37). The kappa (k) coefficient is 0.6413. Table 2 displays the combined distribution of our final corpus.

TABLE II. SUMMARY OF THE CORPUS USED IN THE EXPERIMENTS.

|  | AOL | Sogou |
|---|---|---|
| Navigational | 27,654 | 3700 |
| Transactional | 5415 | 2100 |
| Informational | 26,931 | 4200 |
| Total | 60,000 | 10,000 |

*B. Baseline Models*

We use two basic click models, DBN and UBM and extend them to intent-aware models, to illustrate the impact of our proposed search intent bias. We use the the implementations of the baseline models provided by Chuklin et al. [23] and make some necessary modifications to adapt them for a fair comparison on our dataset. The intent-aware versions of the two models are the orignial ones after being replaced the examination bias represented by formula (6) and (7) with the formula (18) and (19), which is quite simple.

*C. Search Intent Identification Result*

The labeled datasets are then divided randomly and evenly into a training set and a test set. We use the features listed in table 1 to construct the feature vector and Supported Vector Machine (SVM) with competitive performance in multi-class classification to train the classifier [20]. Table 3 lists the query identification results of AOL and Sogou datasets.

TABLE III. SEARCH INTENT IDENTIFICATION RESULT

|  | Accuracy | Recall | F1 score |
|---|---|---|---|
| AOL dataset | 0.84 | 0.87 | 0.85 |
| Sogou dataset | 0.78 | 0.81 | 0.79 |
| Average | 0.81 | 0.84 | 0.82 |

Because of the complexities of Chinese language, the classification accuracy and recall for Sogou dataset are both lower than AOL dataset, but still achieved the F1 score 0.79 which is relatively high. As discussed in subsection 3.2, users show similar click activities when search intents are navigational and transactional. Therefore, one possible method to increase the accuracy and recall is to combine navigational and transactional categories as non-informational category to train a binary classifier, which can be explored in the future.

*D. Click Prediction*

To answer RQ1, we use click perplexity as metric to compare the models based on their fitness to the data. Perplexity can be seen as the log-likelihood powers which are computed independently at each position. For example, we assume that $q_j^s$ is the probability of some click calculated from a click model, i.e. $P(C_j^s = 1)$ where $C_j^s$ is a binary value indicating the click event at position *j* in query session *s*. Then the click perplexity at position *j* is computed as follows:

$$p_j = 2^{-\frac{1}{|S|}\Sigma_{s\epsilon S}(C_j^s log_2 q_j^s + (1-C_j^s)log_2(1-q_j^s))}$$

The perplexity of a data set is defined as the average of perplexities in all positions. Thus, a smaller perplexity value indicates a better consistency between the click model and the actual click data. The improvement of perplexity value p1 over p2 is given by $\frac{p_2-p_1}{p_2-1} * 100\%$ [46]. Table 4 shows the experiment results.

TABLE IV. THE PERPLEXITY COMPARISON OVER RANKING POSITIONS. "@N" REPRESENTS THE PERPLEXITY AT POSITION N. "IMPR." REPRESENTS THE IMPROVEMENTS OF THE INTENT-AWARE (IA) VERSION OVER THE ORIGINAL MODEL.

|  | @1 | @2 | @3 | @4 | @5 | Overall |
|---|---|---|---|---|---|---|
| UBM | 1.771 | 1.310 | 1.202 | 1.088 | 1.083 | 1.268 |
| IA-UBM | 1.731 | 1.295 | 1.197 | 1.086 | 1.082 | 1.255 |
| Impr. | 5.1% | 4.8% | 2.4% | 2.2% | 1.2% | 4.8% |
| DBN | 1.765 | 1.293 | 1.211 | 1.107 | 1.089 | 1.216 |
| IA-DBN | 1.724 | 1.278 | 1.192 | 1.089 | 1.076 | 1.203 |
| Impr. | 5.3% | 5.1% | 9% | 16.8% | 14% | 6% |

We can see that the intent-aware models can achieve 4.8% and 6% improvements over original UBM and DBN respectively. We performed the t-test and it shows that the P-values are both less than 0.01% due to the large-scale of the dataset. We further investigated the perplexity in different ranking positions. It first showed that DBN performs better than UBM on perplexity (the lower, the better). This is consistent with the results reported in [24]. Second, it demonstrates that the intent-aware models achieve improvements over almost all positions. As the ranking position goes lower, the improvement decreases for intent-aware UBM, but increases for intent-aware DBN. The possible reason might be UBM exhibits information transitivity from higher ranking position to lower position, that is different from the DBN model. The UBM model maintains a global matrix, so in prediction, documents in lower positions can benefit from the click/skip information in the higher positions. The lower a position is, the more information it can obtain from higher positions. Therefore, for a document in position like 4 or 5, it may have enough information to calculate the click probability, even though it is very beneficial in the high positions.

*E. Relevance Estimation*

As a click model provides a prediction on the relevance of a document for a query, we can rank documents according to the value of predicted relevance, and compare this predicted ranking with ideal ranking derived from the editorial judgement. We expect the accuracy of the relevance estimation can be improved after eliminating the effect of the search intent bias. The accuracy can be measured by stand IR evaluation metric. In this study, we use Normalized Discounted Cumulative Gain (NDCG) [22] as the evaluation metric for the relevance estimation task. NDCG is calculated cumulatively from the top of the result list to the bottom with the gain of each result discounted at lower ranks. Higher NDCG values correspond to a better ranking result. The NDCG at a particular rank threshold K is defined as:

$$NDCG@K = \frac{1}{Z@K}\sum_{i=1}^{K}\frac{2^{g_i}-1}{\log(1+i)}$$

where Z@K is the normalization to make the ideal ranking (i.e. the ranking obtained by ordering the documents according to their editorial relevance) to have NDCG value of 1, and $g_i$ denotes the editorial relevance of the documents ordered by the ranking. In this paper, we choose two popular models, DBN and UBM and extend them to intent-aware models, to illustrate the impact of our proposed search intent bias. We compared their NDCG scores over queries with three types of search intents and use their arithmetic mean. Experiment results are shown in Fig. 3.

---
[7] https://www.researchgate.net/publication/291184600_AnnotatedCorpus

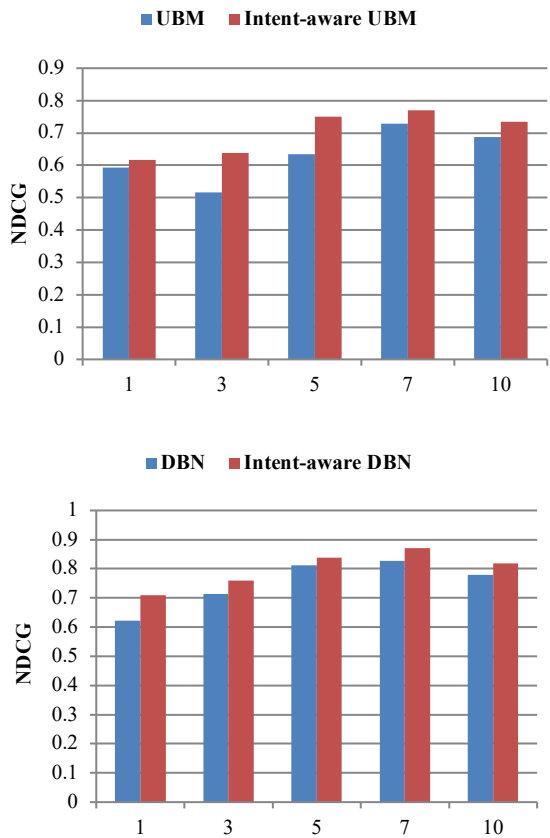

Figure. 3. NDCG values @{1,3,5,7,10} of UBM, DBN and their extended models with intent-aware properties.

From Fig. 3, we clearly see that the new click models extended with the search intent bias hypothesis outperform previous models only with the examination hypothesis. The largest improvement is 23.4% for UBM with NDCG@3. The average NDCG score increases from 0.632 to 0.7 for UBM with an improvement of 11.6%, and 0.75 to 0.8 for DBN with an improvement of 6.7%. This result shows the significance of considering the search intent bias in click models and proves the effectiveness of our proposed models.

## V. CONCLUSION AND FUTURE WORK

In this paper, we study the diversity of user search intent and its influence on user click activities. We found that there exists a bias for clicks among different types of search intents, and thus proposed user search intent bias and rebuild the classical position-based model by adding a search intent labeling variable. Such an extended position-based model can be integrated easily into other complex models. Experiments show that the new models with the search intent hypothesis significantly perform better than the original versions of click models.

Because of the commercial protection of search engine companies, we do not have access to their search logs. In this paper, we used a small open sourced dataset for the experiment, but in the future, we plan to obtain search data by developing a meta search engine and hire students to use it and then collect more search logs. We are also interested in exploring the possibilities of integrating search intent bias into more click models.

ACKNOWLEDGMENT

This work was supported by the Ohio Department of Higher Education, the Ohio Federal Research Network and the Wright State Applied Research Corporation under award WSARC-16-00530 (C4ISR: Human-Centered Big Data).